\documentclass[sn-basic]{sn-jnl}

\usepackage[utf8]{inputenc}
\usepackage{graphicx}
\usepackage{url}
\graphicspath{ {images/} }
\usepackage{caption}
\usepackage{amsmath}
\usepackage{verbatim}
\usepackage{subfig}

\usepackage{adjustbox}
\usepackage{multirow}

\usepackage[referable]{threeparttablex} 
\usepackage{tikz}
\def\checkmark{\tikz\fill[scale=0.4](0,.35) -- (.25,0) -- (1,.7) -- (.25,.15) -- cycle;}

\usepackage[authoryear]{natbib}

\RequirePackage{fix-cm}

\def\tsc#1{\csdef{#1}{\textsc{\lowercase{#1}}\xspace}}
\tsc{WGM}
\tsc{QE}
\tsc{EP}
\tsc{PMS}
\tsc{BEC}
\tsc{DE}

\usepackage{hyperref}
\hypersetup{
    colorlinks=true,
    linkcolor=blue,
    filecolor=magenta,      
    urlcolor=cyan,
    pdftitle={Overleaf Example},
    pdfpagemode=FullScreen,
    }    

\begin{document}

\title[Machine Learning Competition Ecosystem]{The ecosystem of machine learning competitions: Platforms, participants, and their impact on AI development}

% \orcidlink{0000-0002-1765-0646}
\author[1]{\fnm{Ioannis} \sur{Nasios} } 
\email{ioannis.nasios@nodalpoint.com}
\affil[1]{\orgname{Nodalpoint Systems},
\orgaddress{\street{Pireos 205}, \city{Athens}, \postcode{11853},\country{Greece}}}

% \maketitle
% \begin{frontmatter}

% \begin{abstract}
\abstract{
Machine learning competitions (MLCs) play a pivotal role in advancing artificial intelligence (AI) by fostering innovation, skill development, and practical problem-solving. This study provides a comprehensive analysis of major competition platforms such as Kaggle and Zindi, examining their workflows, evaluation methodologies, and reward structures. It further assesses competition quality, participant expertise, and global reach, with particular attention to demographic trends among top-performing competitors. By exploring the motivations of competition hosts, this paper underscores the significant role of MLCs in shaping AI development, promoting collaboration, and driving impactful technological progress.  Furthermore, by combining literature synthesis with platform-level data analysis and practitioner insights a comprehensive understanding of the MLC ecosystem is provided. 

Moreover, the paper demonstrates that MLCs function at the intersection of academic research and industrial application, fostering the exchange of knowledge, data, and practical methodologies across domains. Their strong ties to open-source communities further promote collaboration, reproducibility, and continuous innovation within the broader ML ecosystem. By shaping research priorities, informing industry standards, and enabling large-scale crowdsourced problem-solving, these competitions play a key role in the ongoing evolution of AI. The study provides insights relevant to researchers, practitioners, and competition organizers, and includes an examination of the future trajectory and sustained influence of MLCs on AI development.
}
% \end{abstract}

% \begin{keyword}
\keywords{Global AI Participation; Crowdsourced AI; Competitive Machine Learning;  ML Competition Platforms; Kaggle;}
% \end{keyword}
\maketitle
% \end{frontmatter}

\section{Introduction}
\label{sec:introduction}
%overview
Machine Learning Competitions (MLCs) are essential drivers of Artificial Intelligence (AI) innovation, bridging the gap between academic research, industry applications, and talent development. By analyzing competition formats, reward structures, hosts and participant engagement across multiple platforms, it is demonstrated that MLCs foster skill acquisition, career growth, and real-world AI development. The study highlights Kaggle's dominance in participation volume and prize distribution, the specialized focus of platforms like Drivendata and Zindi, and the impact of ranking systems on professional recognition. Through participation trends, demographic insights, and prize allocation data, concrete evidence of MLCs' role in shaping the Machine Learning (ML) ecosystem is provided, influencing research directions and driving industry-academia collaboration.

Annual reviews by \cite{carlens2025state, carlens2026state} highlight the continued expansion of MLCs, with hundreds of competitions across multiple platforms awarding tens of millions of dollars in prizes. High-stakes “grand challenge” competitions offering prizes above \$1 million have re-emerged, while common winning approaches continue to rely on tools such as Python, PyTorch, and gradient-boosted trees. Recent trends include the use of quantization in LLM-related tasks and ongoing evaluation of AutoML methods. Competitions are increasingly used by corporations, nonprofits, and public organizations to address real-world problems, while academia leverages them for benchmarking state-of-the-art (SOTA) techniques. The ecosystem remains dominated by major platforms such as Kaggle and Tianchi, but is also diversifying through open-source frameworks like Codabench. Overall, MLCs have evolved into important environments for evaluation, knowledge sharing, and the advancement of both applied and research-oriented ML.

MLCs especially on platforms like Kaggle, provide an effective pedagogical framework that blends game-based learning with social constructivism. \citet{chow2019pedagogy} proposed a seven-step competition-based teaching model that enhances engagement, collaboration, and conceptual understanding. Similarly, studies by \citet{kalazhokov2023methodology} and \citet{duran2022gamifying} demonstrated that integrating Kaggle participation into data science curricula fosters strong motivation and accelerates skill acquisition through real-world datasets, instant feedback, and competitive dynamics. These competitions effectively simulate real-world problem-solving, bridging theoretical knowledge with practical application across computing and engineering disciplines. \citet{chang2024applying} further showed that competition-based learning (CBL) strengthens both technical and collaborative abilities, preparing students for innovation and real-world challenges in AI and data science.

% importance of AI for industry
The growing significance of AI within the industrial sector has been extensively examined in recent years. \citet{su2023machine} conducted a comprehensive review compiling diverse ML methods applied to the diagnostics and prognostics of industrial systems, utilizing open-source datasets originating from Prognostics and Health Management (PHM) data challenges. \citet{enholm2022artificial} underscored AI's pivotal role in enhancing business value and competitive advantage through improved operational efficiency and data-driven decision-making. Similarly, \citet{li2024reskilling} framed AI as a transformative force that extends beyond automation, emphasizing the need for organizations to invest in reskilling and upskilling initiatives to prepare their workforce for evolving technological demands. As industries continue to adapt to this transformation, the cultivation of advanced AI and data science skills has become critical for maintaining competitiveness. In this context, MLCs have emerged as powerful enablers of AI education and workforce development, offering accessible, practice-oriented environments that bridge theoretical learning with real-world application.

The convergence of crowdsourcing and data science has deepened the relationship between industry and the global research community. \citet{tauchert2020crowdsourcing} identified over thirty organizations hosting multiple competitions, highlighting the collective problem-solving capacity of the data science community. However, these initiatives also present challenges related to data governance, privacy, and intellectual property. While open competitions stimulate innovation and talent discovery, they demand thoughtful design to ensure ethical integrity and secure data handling. As \citet{hind2025challenges} noted, initiatives like the Waymo Open Dataset Challenges serve as catalysts for research and industrial progress. To maximize their impact, competition organizers must prioritize ethical standards, interdisciplinary collaboration, and rigorous dataset preparation to balance innovation with responsible AI development.

% Competition orginizers
MLCs have played an increasingly prominent role in the development and evaluation of modern techniques, contributing to both academic research and industry practice. These competitions foster research in multiple ways, with various stakeholders leveraging them for different purposes. Competition organizers often conduct research based on the competition structure, dataset, and outcomes, analyzing key trends and advancements in ML methodologies. For example, in the M5 Accuracy competition, the results, findings, and conclusions were synthesized into a comprehensive analysis \citep{makridakis2022m5}, providing valuable insights into forecasting challenges and best practices. Similarly, \cite{suenderhauf2019probabilistic} organized a competition on the Codalab platform to enhance probabilistic object detection, inviting the top four participants to present their findings at a CVPR workshop.

% Author's papers after competitions
Building on the tradition of competition methodologies evolving into academic contributions, prior work shows how participation in MLCs can lead to impactful scientific outcomes. For example, methods developed in the M5 forecasting competition were later formalized for hierarchical time series forecasting \citep{nasios2022blending}, while related approaches have been applied to scientific domains such as mass spectrometry analysis \citep{nasios2024analyze} and landslide detection using multi-source satellite data \citep{nasios2026landslide}. In environmental monitoring, SOTA computer vision and data fusion techniques have been used for tasks such as kelp forest detection and algal bloom classification \citep{nasios2025enhancing, nasios2025ai}. Overall, these studies illustrate how competition-driven experimentation can translate into rigorous academic research and broader scientific advancement.

% important competitions
Among the most influential milestones in the history of MLCs are a few landmark challenges that have significantly shaped the field's trajectory. The Netflix Prize \citep{bennett2007netflix}, launched in 2006 with a \$1 million award, demonstrated the transformative power of open innovation and crowdsourced intelligence by spurring global collaboration to improve movie recommendation algorithms. Similarly, the Data Science Bowl series \citep{kuan2017deep,caicedo2019nucleus}, organized by Booz Allen Hamilton in partnership with Kaggle, has become one of the largest challenge series in applied data science, offering substantial prizes to address pressing problems in fields such as biomedical imaging, ocean health, and disease diagnostics. Another influential series of challenges advancing both the theory and practice of forecasting are the M Competitions, held six times between 1982 and 2022, reflecting the field's evolution from traditional statistical approaches to modern AI-based forecasting methods \citep{makridakis2020m4,makridakis2022m5,MAKRIDAKIS20251315}. A high-impact initiative, the DeepFake Detection Challenge sponsored by Facebook, AWS, and the Partnership on AI, provided a large-scale benchmark dataset and significant monetary incentives to accelerate research on detecting synthetic media and combating misinformation \citep{dolhansky2020deepfake}. These competitions not only advanced technical capabilities but also underscored the value of incentivized, community-driven problem-solving in tackling complex, real-world issues.

% using datasets from competitions or just from MLCP
MLCs also have a broader and more enduring impact on the research community. Their datasets frequently evolve %into widely used benchmarks
for developing and evaluating new ML models. For example, \cite{punn2022bt} leveraged the Kaggle Data Science Bowl 2018 dataset to propose a self-supervised framework for biomedical image segmentation based on Barlow Twins and U-Net architectures. Similarly, \cite{geng2022pruning} analyzed multiple Kaggle datasets to study efficient pruning strategies for convolutional neural networks using filter similarity. In the context of time series forecasting, \cite{godahewa2023setar} utilized diverse competition datasets to evaluate a novel tree-based approach, while \cite{zhang2022pulmonary} demonstrated the continued utility of competition data through large-scale pulmonary disease detection experiments on a Tianchi dataset. Overall, these examples illustrate how competition datasets support ongoing innovation, reproducibility, and collaboration in the ML community. By providing shared resources and standardized evaluation settings, MLCs help bridge academic research and real-world applications, reinforcing a continuous cycle of experimentation and advancement.

%MLC platforms for conferences
Competitions have become an integral component of academic research, with over a hundred now affiliated with major conferences each year. For instance, \href{https://blog.neurips.cc/2024/06/04/neurips-2024-competitions-announced/}{NeurIPS 2024} featured 16 competitions covering topics such as Generative AI, Reinforcement Learning, and Responsible AI. Similarly, robotics conferences like ICRA hosted 12 official tracks in 2023, while leading events including \href{https://cvpr.thecvf.com/?ref=mlcontests}{CVPR}, \href{https://iccv2023.thecvf.com/?ref=mlcontests}{ICCV}, and \href{https://icml.cc/?ref=mlcontests}{ICML} incorporated competition-based workshops, such as the FGVC series at CVPR, which organized seven challenges. Many of these events rely on platforms like CodaLab \citep{pavao2023codalab}, though some, such as Sensorium 2023, are hosted on independent infrastructures.

%ML advancements in various domains
MLCs are often designed to advance scientific progress within specific domains by harnessing collective intelligence and SOTA ML techniques. \citet{knoll2020advancing} exemplified this through the fastMRI challenge, which aimed to improve MR image reconstruction using advanced ML models. Likewise, \citet{johnson2021laboratory} organized a competition based on laboratory simulation data to enhance earthquake forecasting accuracy. The “IceCube-Neutrinos in Deep Ice” challenge \citep{eller2023public} further stimulated innovation in neutrino event reconstruction, yielding more precise and efficient analysis methods. Similarly, the Spacenet challenge series \citep{van2018spacenet} has been instrumental in advancing remote sensing and satellite imagery analysis. These competitions not only accelerate research progress but also encourage interdisciplinary collaboration and real-world problem-solving.

% the kaggle book, MLC is a school for companies staff, career advansment
As the leading platform for MLCs, Kaggle has been instrumental in shaping the modern AI landscape. In a position paper presented at the ICML conference \citep{sculley2025position}, Kaggle argued that competitions represent the gold standard for ensuring empirical rigor when evaluating Generative AI systems. \citet{banachewicz2022kaggle} emphasized its extensive resources and transformative impact on both participants and organizations. Through active competition, individuals gain practical experience, refine technical expertise, and often leverage their achievements to advance their careers, while companies use such platforms to source innovative solutions. Complementing this perspective, \citet{pavao2023methodology} examined competition design and analysis, suggesting that such community-driven engagement fosters reproducible research and contributes to the democratization of AI. 

% AI competitions platforms direct the AI development and define the generalisability of models
\citet{luitse2024ai}, analyzing 118 competitions hosted on Kaggle and Grand Challenge platforms, showed that MLC platforms shape power dynamics in medical imaging by influencing who can host competitions, how they are structured, and how datasets and generalization criteria are defined. Participants further affect outcomes through their backgrounds, which can limit broader applicability. More generally, competition structures influence model development. \citet{ginart2021competing} found that competitive settings often drive models to specialize in sub-populations at the expense of generalization. While competitions promote innovation and optimization, they may also introduce biases that reduce robustness in real-world applications, highlighting both their strengths and limitations.

% social network importance for a user to succeed in MLCPs
Effective collaboration and competition design play important roles in MLC performance and engagement. \citet{twyman2023positioning} showed that network positioning, collaboration diversity, and connectivity positively influence success in Kaggle competitions. Complementing this, \citet{liu2023designing} argued that multi-stage formats and prize allocation can enhance participant engagement, with two-stage and elimination-based designs helping sustain effort. However, widely used single-stage formats, such as those on Kaggle, can also maintain engagement when paired with clear objectives, transparent evaluation, and active community interaction.

% % Competition results are true while most research finding from non competitions are false
Competition results can provide useful empirical evidence, particularly in large-scale settings with diverse participation. In high-stakes MLCs, curated datasets and broad participation encourage extensive experimentation and validation, although similar rigor is expected in academic research. For example, \citet{roelofs2019meta} found that public and final leaderboard (LB) scores in Kaggle competitions were generally consistent, suggesting a focus on generalization rather than overfitting. At the same time, concerns about research reliability remain, as highlighted by \citet{ioannidis2005most}. In this context, MLCs can complement traditional research by enabling repeated evaluation on shared benchmarks, supporting reproducibility and comparative assessment across methods.

% SOTA claimed in research most of the times will NOT improve performance. Models from Google and META and prior relevant competitions will improve performance.
From practical experience, competitors often consult SOTA research during the competition phase to identify best practices, model architectures, and domain-specific insights. However, many findings reported in academic literature do not consistently translate into performance improvement in competitive settings. In practice, approaches that tend to be more effective include well-established SOTA models released by major industry actors such as Google and Meta, as well as techniques and lessons derived from previous competitions. Prior winning solutions, shared code, and community insights frequently play a significant role in guiding model development and improving performance.

% review's elements
This study combines multiple data sources and methodological approaches to provide an overview of MLCs and their role within the AI ecosystem. A bibliographic review situates the work within existing research on innovation, collaboration, and industry impact. Data from official MLC platforms, the Meta Kaggle dataset, and academic databases are used to examine participation patterns, performance distributions, ranking systems, and related publications. %Web analytics are further considered to assess platform visibility and engagement. 
Additional insights are informed by practical experience with competition platforms. The data sources and methods are described in detail in the following sections to ensure transparency and reproducibility.

% \section{Material and methods}
% \label{sec:Material and methods}
\section{Overview of machine learning competition platforms}
\label{sec:MLCP_overview}
This section examines the major platforms that host MLCs, emphasizing their defining characteristics and ecosystem roles. The analysis combines insights from platform documentation, public data sources, and the author's extensive firsthand experience. This combination of empirical evidence and practitioner perspective enables a nuanced understanding of factors such as prize distribution mechanisms, community engagement, and strategic dynamics influencing success.

\subsection{Major platforms}
\label{sec:MajorPlatforms} 
The following subsection outlines ten of the most prominent MLCPs, presented in alphabetical order. The overview draws on information from official platform sources, complemented by insights derived from extensive direct participation and long-term engagement within these ecosystems.

\subsubsection{Aicrowd}
\label{sec:Aicrowd} 
% \href{https://www.aicrowd.com/}{Aicrowd} 
\cite{aicrowd} is a platform that streamlines AI workflows by hosting challenges for businesses, universities, governments, and NGOs. It connects complex problems with a global data science community, enabling collaborative solution development. AIcrowd supports code submission, evaluation, and versioning, accelerating AI prototype deployment. Originally linked to EPFL's Digital Epidemiology Lab, it has grown into an independent entity, facilitating competitions for various organizations, including well-known names such as Amazon, Meta, and Sony.

\subsubsection{Biendata}
\label{sec:Biendata} 
% \href{https://www.biendata.xyz}{Biendata} 
\cite{biendata} is a data science competition platform that connects organizations with data scientists to tackle real-world challenges. Focused on fostering innovation, Biendata hosts competitions in areas like healthcare, finance, and technology, allowing participants to solve predictive modeling and data analysis problems. The platform offers opportunities for individuals and teams to showcase their skills, win prizes, and gain recognition in the data science community. Biendata hosted the KDD Cups of 2024, 2020, 2018, and 2015, highlighting the platform's involvement in major MLCs.

\subsubsection{Codalab-Codabench}
\label{sec:Codalab} 
\href{https://www.codabench.org/}{Codabench} \citep{codabench} represents the latest stage in the evolution of \href{https://codalab.lisn.upsaclay.fr/}{Codalab}, providing a more flexible and user-friendly framework for organizing AI benchmarks. It enables organizers to define tasks, customize evaluation protocols, integrate compute resources, and support both traditional and inverted benchmarks, while offering participants real-time feedback through LB-based evaluation.
Its predecessor, CodaLab, originally developed in 2013 through a collaboration between Microsoft and Stanford University and later maintained by Université Paris-Saclay, established itself as a widely used open-source platform for scientific competitions, particularly in conference and workshop settings. Building on this foundation, Codabench extends these capabilities, reflecting a broader shift toward more flexible, reproducible, and scalable benchmarking infrastructures in ML.

\subsubsection{Drivendata}
\label{sec:Drivendata} 
% \href{https://www.drivendata.org}{Drivendata} 
\cite{drivendata} is a platform that applies data science and crowdsourcing to address social challenges. Through online competitions brings together a global community of data scientists to solve real-world problems with measurable impact. These competitions focus on issues like healthcare, education, and environmental sustainability. The platform then helps organizations integrate the winning solutions into their operations, enabling more effective and sustainable strategies. By combining cutting-edge ML techniques and open innovation, Drivendata empowers organizations to leverage data in transformative ways, driving positive social outcomes. NASA, NOAA, Microsoft, and Meta have many times hosted competitions on this platform.

\subsubsection{Kaggle}
\label{sec:Kaggle} 
% \href{https://www.kaggle.com}{Kaggle} 
\cite{kaggle} is the leading platform for data science and ML, offering a dynamic environment for learning, collaborating, and competing. It hosts a variety of challenges, from beginner-friendly tasks to high-stakes industry-sponsored competitions, enabling participants to tackle real-world problems, gain hands-on experience, and compete for prizes. Kaggle provides free hosting for datasets in both private and public modes, an interactive coding environment through Notebooks, and the Kaggle Learn platform, which offers structured tutorials on data analysis, ML, and deep learning.

Kaggle competitions come in different formats to cater to all skill levels. Featured competitions attract top experts with significant prize pools, while community competitions allow users to create and participate in unique challenges, without points and medals and usually without monetary rewards. Getting started competitions provide guided, beginner-friendly experiences, whereas playground competitions offer fun, low-stakes problems for intermediate users. Additionally, Kaggle supports different competition structures, including file submission and code competitions.

A key aspect of Kaggle's ecosystem is its points and medals system, originally introduced for competitions but later expanded to Notebooks, Discussions, and Datasets. This gamified approach has successfully fostered knowledge-sharing and increased community engagement. Furthermore, Kaggle provides free access to cloud-based computing resources, including GPUs and TPUs, helping to democratize AI development and support learning across diverse backgrounds. Acquired by Google in 2017, Kaggle is a major force of the ML community, driving innovation and collaboration on a global scale.

\subsubsection{Solafune}
\label{sec:Solafune} 
% \href{https://solafune.com}{Solafune} 
\cite{solafune} is an emerging AI platform designed to support data science and AI enthusiasts through competitions and collaborative opportunities, with a strong focus on leveraging satellite and geospatial data to address global challenges. By providing curated datasets, interactive tools, and educational resources, Solafune enables participants of all skill levels to tackle industry-relevant problems in areas such as environmental monitoring and resource management. Featuring live LBs and recognition for outstanding achievements, Solafune empowers participants to apply cutting-edge data science techniques while driving positive social impact.

\subsubsection{Thinkonward}
\label{sec:ThinkOnward} 
% \href{https://thinkonward.com/app/c/challenges/}{Thinkonward} 
\cite{thinkonward} is a platform designed to accelerate progress in the energy and natural resources sectors by leveraging open collaboration and advanced technology. Following the transition from Xeek.ai, Thinkonward has established strong ties with Shell, focusing on subsurface exploration, carbon capture, and other critical energy challenges. The platform connects energy companies with a global network of geoscientists, data scientists, and engineers to collaboratively solve complex problems and drive innovation. Through a digital ecosystem that supports projects, challenges, and bounties, the platform fosters an environment of collective wisdom and experimentation, driving breakthroughs in energy exploration and technology adoption.

\subsubsection{Tianchi}
\label{sec:Tianchi} 
 % Alibaba's \href{https://tianchi.aliyun.com/}{Tianchi}
 \cite{tianchi} platform is a cloud-based hub for ML and big data innovation, offering real-world datasets to researchers, academics, and businesses. Powered by Aliyun's ODPS, Tianchi promotes crowd intelligence and innovation through big data competitions, a data lab, and academic collaborations. Its competitions challenge participants worldwide to tackle business and social problems, while its curriculum cooperation program supports universities by integrating data mining and ML education with practical experience.
 
\subsubsection{Topcoder}
\label{sec:Topcoder} 
% \href{https://www.topcoder.com/challenges}{Topcoder}
\cite{topcoder} is a competitive programming and AI development platform that connects skilled professionals with companies seeking innovative solutions. It hosts a range of challenges, including algorithmic problem-solving, ML, and data analysis, alongside software and design tasks. With a gamified approach offering cash prizes and career opportunities, Topcoder attracts talents and fosters continuous skill development. Participants engage in single-round matches, marathon competitions, and client-driven challenges, using the platform's collaborative environment to refine their expertise. Notably, Topcoder has hosted the Spacenet MLCs, advancing open-source geospatial research.

\subsubsection{Zindi}
\label{sec:Zindi} 
% \href{https://zindi.africa}{Zindi} 
\cite{zindi} is Africa's first data science competition platform, designed to connect data scientists with organizations seeking data-driven solutions to the continent's most pressing challenges. The platform hosts a vibrant ecosystem of professionals, including scientists, engineers, NGOs, governments, and companies, all focused on addressing social, economic, and environmental issues in Africa. Zindi works with various entities to develop challenges and offers data scientists, from beginners to experts, a space to access African datasets, hone their skills, and showcase their talents. With an emphasis on affordability and creativity, Zindi enables organizations to solve complex problems while fostering the growth of the data science community across Africa.

\subsubsection{Other MLCPs}
\label{sec:other_mlcp} 
Several other MLCPs are noteworthy for their unique contributions and specialized focus areas. A list of some of them is provided below.
\begin{itemize}
\item{} \href{https://bitgrit.net/competition/}{Bitgrit} – A platform that connects data scientists worldwide through competitions, often with an emphasis on AI-driven business solutions and industry applications.
\item{} \href{https://crunchdao.com/?ref=mlcontests}{Crunchdao} – A decentralized research community and platform that connects a global network of ML engineers with institutions to solve complex predictive challenges in high-stakes tasks in fields like that of quantitative finance.
\item{} \href{https://eval.ai/web/challenges/list}{Eval.ai} – A popular choice for hosting MLCs, particularly in academic and research-oriented contexts. It provides a framework for benchmarking models and is widely used in reinforcement learning and natural language processing (NLP) challenges.
\item{} \href{https://grand-challenge.org/challenges/?ref=mlcontests}{Grand-Challenge} – Specializes in medical imaging and healthcare-related competitions. It is a key hub for AI researchers looking to advance medical diagnostics through computer vision and deep learning models.
\item{} \href{https://huggingface.co/competitions?ref=mlcontests}{Hugging Face} – A relatively new but growing MLCP, leveraging its strong open-source community in NLP and transformer-based models.
\item{} \href{https://signate.jp/?ref=mlcontests}{Signate} – A Japan-based ML competition platform with a strong presence in Asia, frequently collaborating with government agencies and corporations to solve real-world business and societal challenges.
\end{itemize}

These platforms, while not as widely recognized as Kaggle or Zindi, offer specialized opportunities for ML practitioners and researchers, catering to niche domains such as healthcare, NLP, and industry-specific AI solutions.

% \subsection{From Kaggle to Global Expansion: The Evolution of MLCPs}
% \label{sec:MLCP_evolution} 
% \subsection{From Kaggle to Global Expansion}
\subsection{Evolution of ML competition platforms}
\label{sec:fromKaggle2Global} 
As shown in \autoref{table:mlcP}, the leading MLCPs are primarily based in the United States, while the rest are distributed globally. This concentration in the U.S. is unsurprising, given the country's significant investments in AI, strong economy, and outward-facing innovation approach. While Topcoder appears to be the oldest platform, it is more of a general coding competition site rather than a dedicated MLCP. The real emergence of MLCPs began with Kaggle in 2010, followed by a first wave of expansion between 2013 and 2015 with platforms like Codalab, Drivendata, Tianchi, and Biendata. A second wave of growth occurred between 2018 and 2020 with the addition of Zindi, AIcrowd, Thinkonward, and Solafune. Kaggle remains the most dominant, hosting competitions across various domains, while other platforms often focus on specific areas. Drivendata emphasizes competitions for social good, including collaborations with prestigious organizations like NASA. Zindi targets challenges benefiting African nations, Thinkonward focuses on the energy and geoscience sectors, and Solafune specializes in earth observation tasks, often leveraging Copernicus Sentinel imagery.

\begin{table*}[ht!]
\caption{Major Machine Learning Competition Platforms}
\label{table:mlcP}
\centering
\begin{tabular}{l  l  l  l} 
 Platform & Headquarters  & Focusing Sector & Found \\
 \hline
 \href{https://www.aicrowd.com/}{Aicrowd}   & Switzerland & Various & 2018 \\
 \href{https://www.biendata.xyz}{Biendata} & China & Various  & 2015 \\
\href{https://codalab.lisn.upsaclay.fr/}{Codalab} -
\href{https://www.codabench.org/}{Codabench} & France & Science & 2013 \\
\href{https://www.drivendata.org}{Drivendata} & USA & Social Good & 2014 \\
\href{https://www.kaggle.com}{Kaggle} & USA & Various & 2010 \\
\href{https://solafune.com}{Solafune} & Japan & Earth Observation & 2020 \\
\href{https://thinkonward.com/app/c/challenges/}{Thinkonward}   & USA & Energy \& Geoscience & 2020 \\
\href{https://tianchi.aliyun.com/}{Tianchi}   & China & Various & 2014 \\
\href{https://www.topcoder.com/challenges}{Topcoder} & USA - Global & Coding & 2001 \\
\href{https://zindi.africa}{Zindi}  & South Africa & Helping Africa & 2018 \\

% \multicolumn{5}{l}\footnotesize{xyz}\\
% \hline
\end{tabular}
\end{table*}

% \subsection{Competition Workflow and Evaluation Process}
\subsection{Competition workflow and evaluation}
\label{sec:comp_workflow} 
Participating in a MLC follows a structured process designed to ensure fairness and transparency. Competitors begin by creating an account on the platform (if they do not already have one) and signing in. They must then accept the competition's terms and conditions before gaining access to the dataset. After analyzing the data and developing their models, participants generate and submit either their output predictions or the code required to produce them. Typically, competitors are allowed multiple submissions but must select a limited number, usually two, as their final entries. At the conclusion of the competition, the private LB is revealed, showing the provisional rankings based on hidden test data. The top-ranked participants are then required to submit additional documentation and their code for verification. Organizers review these submissions, disqualify any cheaters, and validate the final results before confirming the final LB. Once the winning solutions are officially approved, the prize money is transferred to the respective winners.

Cash prizes are typically awarded to the top (usually three) positions of the final LB, with the amount varying based on rank. The method of prize distribution differs across platforms, as each prefers a specific approach. While direct bank transfers to the winner's account are common, some platforms opt for payment services such as Payoneer, Wise, or Tipalti. Regardless of the method, tax obligations fall on the recipient rather than the platform. Furthermore, for platforms based in the U.S., winners must complete a specific tax form before receiving their payout. Whether taxes are withheld depends on the tax agreement between the U.S. and the winner's country of residence.

% \section{MLCPs Formats, Participation, and Influence}
\section{Structure, participation, and influence of MLCPs}
\label{sec:MLCPs_FPI} 
MLCPs provide AI practitioners with opportunities to test their skills, collaborate, and drive innovation. These platforms differ in competition formats, reward structures, and participant engagement, influencing both the quality of challenges and competitor expertise. Examining their structures, prize distributions, and ranking systems offers key insights into their impact on AI research, industry adoption, and professional development. 

% \subsection{MLCPs formats and reward systems}
% \subsection{Formats and reward systems}
\subsection{Competition formats and reward systems}
\label{sec:formatreward} 
Analyzing the differences among MLCPs provides valuable insights into their effectiveness in fostering innovation and user engagement. Platforms vary in competition formats and reward systems. In contrast to the one-stage competition structure as described in \autoref{sec:comp_workflow}, some platforms adopt a two-stage competition structure, beginning with a development phase where participants submit results using a validation set, followed by a test phase during which submissions are performed using a newly released test set. In addition to cash prizes awarded to top-ranking participants, many platforms offer rewards such as points and medals, which contribute to the participant's profile. These points accumulate over time, updating the global ranking of platform users, while medals are typically awarded based on percentile rankings. For example, on Kaggle, participants in the top 10 plus 1 for every 500 competition participants on the final LB receive a gold medal, the top 5\% earn silver medals, and the top 10\% receive bronze medals, all of which are added on their profile pages as indicators of their ML expertise.

\autoref{table:compchar} summarizes the key attributes of the examined platforms. Platforms employing a two-stage evaluation format typically do not maintain a global ranking or medal system, as fair reward allocation would be difficult for participants who only complete the first stage. Among one-stage evaluation platforms, Kaggle and Zindi both feature comprehensive medal and point systems. Solafune also offers medals and rankings, however global LB updates occur infrequently and are steep, likely due to the small number of competitions hosted annually. Thinkonward similarly provides points and medals, though the limited participant pool in each competition results in few medals being awarded, with only the top ten global rankings publicly displayed. In contrast, DrivenData does not utilize either a medal or point system, possibly reflecting participation constraints in certain competitions permitted to U.S. residents only.

\begin{table*}[ht!]
\caption{MLCP attributes}
\label{table:compchar}
\centering
\begin{tabular}{l | c c c c} 
 Platform & Stages & Ranking & Medals & Hosts\\
 \hline
Aicrowd & 2 &  &  & Various \\ 
Biendata & 2 &  & & Various  \\ 
Codalab/bench & 2 &  & & Various + Open Source \\ 
Drivendata & 1 &  & & Various \\ 
Kaggle & 1  & \checkmark & \checkmark & Various + Community \\ 
Solafune & 1 & \checkmark & \checkmark & Self \\
Thinkonward & 1 & \checkmark & \checkmark & Self \\
Tianchi & 2 &  &  & Various \\
Topcoder & 2 &  &  & Various \\ 
Zindi  & 1 & \checkmark & \checkmark & Various \\ 
   
\end{tabular}
\end{table*}

\subsection{Competition hosts}
\label{sec:competition_hosts} 
A crucial element of MLCs, beyond the platforms and participants, is the competition hosts. Hosts range from individual researchers and academic institutions to companies, NGOs, and government agencies, each bringing unique objectives and expertise. The diversity of competition hosts contributes to the broad applicability of MLCs across industries and disciplines, driving AI innovation in both commercial and non-commercial domains.

As presented in \autoref{table:compchar}, Solafune and Thinkonward probably host their own competitions rather than facilitating challenges from external organizers. In contrast, all other platforms accommodate competitions from various sources. Kaggle stands out among MLCPs in its hosting model, as it not only supports competitions from companies and institutions but also allows individuals to create their own contests for free in the community section. However, these community-hosted competitions typically receive less attention since they do not award medals or ranking points. Furthermore, Codalab offers full flexibility, enabling anyone to host a competition. As it is an open-source project, it can also be installed on private servers or run locally, providing a customizable alternative for hosting AI challenges.

Various organizations, from tech giants to research institutions and nonprofits, actively fund and host MLCs across different platforms. Many competitions on the Zindi platform have been funded by Microsoft, Amazon, AI for Good and UNICEF. Kaggle has hosted many competitions for Kaggle itself, other Google entities as Google research or Google Brain, various universities, laboratories, foundations, companies and even for some competitors such as Facebook and Microsoft. Drivendata has hosted many competitions for major entities as Meta, Microsoft, NASA, NOAA and other agencies. Codalab is increasingly favored for hosting competitions for conferences and at workshops while AIcrowd has hosted many competitions for the famous NeurIPS conference, and lately for Meta and Amazon. 

AI remains a highly dynamic field, with leading tech companies investing heavily across various domains. Google, through Kaggle, has established itself as the dominant force in competitive ML, making it challenging for any competitor to rival its position. While Microsoft, Amazon, and Meta do not operate dedicated MLCPs, they have actively participated by hosting numerous competitions on other platforms. Their strong presence in the competitive ML landscape underscores the growing significance of MLCs in driving innovation and industry engagement.

MLCs linked to annual conferences and potential publications are gaining traction, offering unique problem sets that appeal to both advanced participants and college students. Similarly, prestigious annual competitions, such as the renowned ``KDD Cup'', continue to attract significant interest. Achieving a top placement in such a competition can serve as both a challenge and a prestigious accomplishment for ML practitioners, further motivating engagement in these high-profile events.

The motivations behind creating MLCs vary widely depending on the host.  While academic competitions are typically driven by the pursuit of knowledge, industry-led challenges often prioritize profit-driven innovation. Businesses often use competitions to crowdsource innovative solutions, identify top talent, or enhance their brand visibility. Academic institutions leverage them for research advancements, benchmarking, and educational purposes. NGOs and government organizations may focus on addressing critical societal challenges, such as climate change or public health. Platforms themselves sometimes organize competitions to engage their communities and showcase emerging technologies. As MLCs serve as powerful tools for recruitment, networking, data utilization, and publication opportunities, their importance is underscored across different sectors.

% \subsection{Assessing Platform Influence: Community Engagement and Prize Distribution}
% \subsection{Community Engagement and Prize Distribution}
\subsection{Community engagement and prizes}
\label{sec:platformInfluence} 

The significance of a MLCP in the field is determined by several key factors, with the most crucial being the size and quality of its community, the participants engaging in competitions. While the number of registered users in the platform is often used as a measure, it does not accurately reflect actual engagement, as some users may register without participating or may had only competed briefly. A more meaningful metric is the average number of participants per competition within a year, providing a clearer representation of the platform's active user base.

\autoref{table:mlcC} and \autoref{table:mlcC25} provides an overview of key competition metrics across platforms, including the average number of teams per competition, the total number of competitions hosted, and the total monetary prizes awarded in 2024 and 2025 respectively. Notably, only competitions offering monetary prizes were considered in these calculations, although many additional contests took place without offering cash rewards. Furthermore, some competitions advertise larger prize pools, but these are often contingent on achieving specific performance thresholds. Therefore, the prizes reported here correspond only to amounts that were actually awarded. %Furthermore, some competitions may claim of awarding bigger prizes, this sometimes is related to the performance and the majority of the prize is released only when a certain threshold is surpassed. Therefor, Listed prizes here are those that have been awarded.

\begin{table*}[ht!]
\caption{Platform competitions in 2024}
\label{table:mlcC}
\centering
\begin{tabular}{l | r c  r r} 
 Platform & Teams & Competitions & Prizes (\$) & Prize per Team\\
 \hline
Aicrowd & 83 & 11 & 123.000 &  135\\ 
Biendata & 61 & 7 &   55.000 &  129\\ 
Codalab & 26 & 10 &   93.000 &  358\\ 
Drivendata & 73 & 11 &650.000 &  809\\ 
Kaggle & 1.869 & 28 & 2.380.000 &  45\\ 
Solafune & 111 & 4 &  46.000 &  104\\
Thinkonward  & 45 & 11 & 320.000 & 649 \\
Tianchi & 131 & 14 &   65.000 & 35 \\
Topcoder & 45 & 5 &   29.000 & 129 \\ 
Zindi  & 123 & 31 &   147.000 & 39 \\ 

% \multicolumn{6}{l}\footnotesize{xyz} \\
% \hline
\end{tabular}

\end{table*}

\begin{table*}[ht!]
\caption{Platform competitions in 2025}
\label{table:mlcC25}
\centering
\begin{tabular}{l | r c  r r} 
 Platform & Teams & Competitions & Prizes (\$) & Prize per Team\\
 \hline
Aicrowd & 92 & 5 & 118.000 &  257\\ 
Biendata & NA & 11 & 112.000 &  NA\\ 
Codalab & 153 & 37 & 163.000 &  29\\ 
Drivendata & 45 & 3 & 360.000 &  2667\\ 
Kaggle & 1457 & 34 & 3.718.000 &  75\\ 
Solafune & 99 & 2 &  24.000 &  121\\
Thinkonward  & 67 & 4 & 100.000 & 373 \\
Tianchi & 777 & 45 & 67.000 & 2 \\
Topcoder & 128 & 1 &   2.280 & 18 \\ 
Zindi  & 198 & 28 &   146.200 & 26 \\ 

% \multicolumn{6}{l}\footnotesize{xyz} \\
% \hline
\end{tabular}

\end{table*}

The Prize per Team metric, calculated as Prize / (Teams × Competitions), provides an estimate of potential earnings per participant. Based on this measure, platforms such as Drivendata and Thinkonward appear to offer comparatively higher prize opportunities per team, while platforms like Kaggle, Tianchi, and Zindi show lower values, reflecting higher participation levels.

In terms of overall prize distribution and participation, Kaggle stands out as the most prominent platform, exceeding others by a considerable margin. Across the 10 MLCPs examined, approximately \$4–5 million USD in competition prizes were awarded annually over the past two years. Kaggle accounted for about 70\% of the total prize pool, while U.S.-based platforms collectively contributed close to 90\%, indicating a strong concentration of competition funding in the United States.

% \subsection{Evaluating Competition Quality and Participant Expertise}
\subsection{Evaluation of competition quality and participant expertise}
\label{sec:CompetitionQuality} 
Beyond platform's community size, the quality of its community is also crucial. A diverse participant base, including beginners, mentors, and elite competitors, shapes the overall competitiveness and impact of the platform. Highly skilled participants elevate the standard of final solutions, while a strong foundation of learners enhances the platform's role as an educational hub, further solidifying its influence in the ML ecosystem.

Kaggle, the leading MLCP, is unique in assigning tiered rankings, such those of Master and Grandmaster, which can serve as indicators of participant expertise. These rankings can also provide insight into competitor quality beyond Kaggle, offering a reference point for evaluating competitions on other platforms too. Many competitors engage in multiple platforms, driven by various motivations, such as better chances of winning cash prizes due to lower competition density, personal interest in specialized domains, or a desire to gain a well-rounded perspective on ML challenges.

Another way to assess both participant and competition quality is by analyzing the number of competitions in which the contestant has secured a top-3 or prize-ranking position across different MLCPs. However, tracking contestants across platforms can be challenging and often imprecise. Nevertheless, when combined with other factors of the competition, such as total prize value, the number of competitors, and the credibility of the hosting organization, such data can offer major insights into assessing the competition quality.

Conversely, the geographical diversity of participants can also reflect a platform's competitive level. Some platforms, such as Biendata, Solafune, and Tianchi, tend to attract primarily domestic participants rather than a global audience. This limited reach may indicate a lower overall quality of competition compared to platforms with broader international participation, highlighting the platform's importance in estimating the level of a competition.

% % \subsection{Statistical information of competitors in Kaggle}
% % \subsection{Statistics and Demographics of Kaggle's Top Competitors}
% % \subsection{Demographics and Performance Trends Among Kaggle Users}
\subsection{Kaggle user demographics and performance patterns}

\label{sec:stat_kaggle} 
Kaggle has made its metadata file publicly available \citep{megan_risdal_timo_bozsolik_2022}, granting open access to user-related information and other relevant data. By analyzing this dataset to gain insights into participant activity up until the end of 2025, as explored in \url{https://github.com/IoannisNasios/various/blob/master/c_m_gm_g_h_.ipynb}, %\href{https://www.kaggle.com/code/ouranos/c-m-gm-g-h}{https://www.kaggle.com/code/ouranos/c-m-gm-g-h}, 
several key characteristics have been identified. The summarized \autoref{table:mgmu}, offers an understanding of the engagement and achievements of Kaggle users.

\begin{table*}[ht!]
\caption{Kaggle users until the end of 2025}
\label{table:mgmu}
\centering
\begin{tabular}{l r  } 
 Characteristic & Amount  \\
 \hline
Total registered users & 30.387.257 \\ 
Users registered over the last 5 years (2020-2025) & 24.773.771  \\ 
Number of users in rank 1 & 22 \\
Number of competition Grandmasters & 387 \\ 
Mean yearly new Grandmasters (2020-2025) & 35.5 \\ 
Number of competition Masters & 2209 \\
Mean yearly new Masters (2020-2025) & 172.8 \\ 

% \hline
\end{tabular}
\end{table*}

\begin{figure}[ht!]
    \centering
    \includegraphics[width=0.5\textwidth,trim=0 0 0 0, clip]{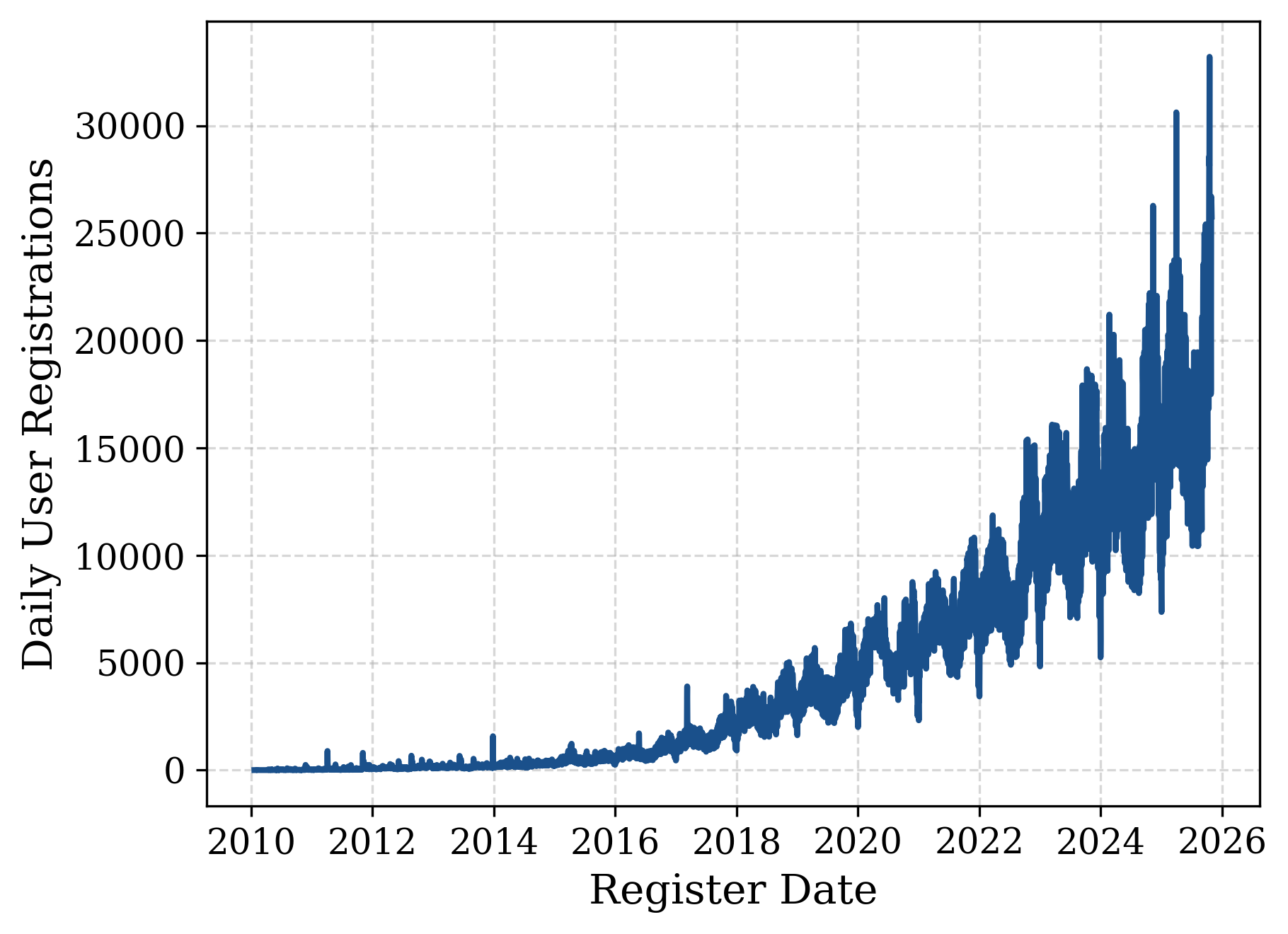}
    \caption{Kaggle daily sign-ups}
    \label{fig:kaggle_registrations}
\end{figure}

\autoref{fig:kaggle_registrations} illustrates a notable exponential growth in the number of registered users on Kaggle. This growth becomes particularly pronounced after 2017, coinciding with Google's acquisition of Kaggle, which likely contributed to the platform's increased visibility and resources. Additionally, the data reveals a recurring annual pattern in user registrations, with a noticeable dip in new sign-ups during the summer months. This seasonal variation could be attributed to factors such as vacation periods, academic cycles, or other external influences that impact user engagement during those months.

Kaggle's global LB incorporates a point decay mechanism to ensure that rankings reflect both current performance and past achievements. This system not only maintains the LB's relevance but also encourages ongoing participation by motivating competitors to remain active. As a result, the top-ranked position has changed multiple times over the years. By the end of 2025, a total of 22 users had reached the number-one rank, with one holding the title of Competition Master and 21 achieving the Grandmaster status.

As indicated in \autoref{table:gmmg}, Japan leads both in the number of Grandmasters and in total gold medals earned by Grandmasters. Although its overall user base is smaller than that of other leading countries, this may reflect a high level of engagement among top participants and a strong focus on achieving top performance.

\begin{table*}[ht!]
\caption{Grandmasters (GMs), Masters and Gold Medals (GM)}
\label{table:gmmg}
\centering
\begin{tabular}{l | r r r r r r r} 
 Country	& GMs & GMs GM & Masters & M. GM & Both & All GM & Users \\ 
 \hline

Japan & 84 & 659 & 303 & 513 & 1.172 & 1.293 & 25.661 \\ 
U.S.A. & 65 & 554 & 393 & 667 & 1.221 & 1.607 & 395.089 \\ 
China & 44 & 369 & 249 & 449 & 818 & 1.112 & 52.061 \\ 
Russia & 16 & 103 & 82 & 120 & 223 & 289 & 32.327 \\ 
U.K. & 14	& 106 & 73 & 132 & 238 & 292 &  40.497 \\ 
India & 8 & 56 & 43 & 77 & 133 & 182 & 282.294 \\ 
Germany & 7 & 96 & 71 & 126 & 222 & 301 & 25.731 \\ 
Australia & 7 & 58 & 30 & 57 & 115 & 146 &  21.229 \\  
Canada & 7 & 58 & 41 & 77 & 135 & 169 & 34.463 \\ 
France & 6 & 62 & 60 & 104 & 166 & 215 & 24.205 \\ 
   
\end{tabular}
\end{table*}

Although EU countries are typically not treated as a single entity, aggregating their statistics allows for a meaningful comparison with larger in population countries such as the USA, China, and India. By aggregating data from the 27 EU member countries, it is found that there are 49 Competition Grandmasters who have collectively earned 495 gold medals, alongside 349 Masters with 599 gold medals. In total, Grandmasters and Masters have accumulated 1,194 medals, while the overall number of medals awarded to EU competitors stands at 1,400, with a total of 148,541 registered users. Based on these, the EU ranks third, following the USA. Interestingly, Germany, EU's member with the highest rank in the table, despite earning a substantial number of gold medals, has relatively fewer Grandmasters, possibly indicating a lower emphasis on acquiring Kaggle titles.

India is rapidly emerging as a dominant force in ML competitions, adding 139,446 new users in the past five years, far surpassing the USA (81,356 new users), with Brazil, Pakistan and China following at around 17,000 new users each. Other countries experiencing notable growth in Kaggle participation include Indonesia, Bangladesh, and Nigeria. Finally, Australia stands out for its relatively high number of Grandmasters compared to its number of Masters and gold medals.

\subsection{Kaggle's ranking system: Motivation, prestige, and industry impact}
\label{sec:ranking_importnace} 
% \subsection{The importance of the global ranking system and tiers}
% \label{sec:ranking_importnace} 
 
Kaggle has implemented a global ranking system since it's early days, through which users accumulate points and earn medals based on competition performance, contributions to discussions, and code sharing. Higher placement in competitions results in more points, while achieving specific ranks awards gold, silver, or bronze medals. This system offers a measurable indication of user skill, time investment, and commitment, while also encouraging continued participation across multiple competitions. Developing a strong portfolio on the platform can serve as a useful signal of applied ML skills, and in some cases has been associated with professional opportunities. There are documented examples of high-performing participants receiving job offers or interview opportunities. However, systematic evidence on the extent to which such portfolios directly translate into employment outcomes remains limited, and they are typically considered alongside broader experience and skills. In addition to the global ranking, Kaggle classifies users into tiers, Grandmasters, Masters, Experts, Contributors, and Novices, based on medal achievements. The effort required to attain higher tiers motivates sustained engagement and serves as an indicator of a user's expertise and competence in the field.

Several companies have strategically built elite Kaggle teams, employing some of the world's top Kaggle Grandmasters to showcase expertise, promote their software and services, and strengthen their presence in the competitive ML landscape. Notably:
\begin{itemize}
\item \href{https://h2o.ai/company/team/kaggle-grandmasters/}{H2O} has assembled a team of 7 Competition Grandmasters who have collectively earned 124 gold medals.
\item \href{https://www.nvidia.com/en-eu/ai-data-science/kaggle-grandmasters/}{Nvidia} employs 15 Grandmasters with 327 gold medals.
\item \href{https://www.rist.co.jp/en/kaggle/}{Rist} has 9 Grandmasters with 113 gold medals.%, plus 1 additional Grandmaster as an advisor with 15 gold medals.
\end{itemize}

The value of excellence in competitive ML is further underscored by the fact that one Grandmaster from H2O and Rist, and two from Nvidia, have reached Kaggle's \#1 rank at different points in time. These professionals continue to participate in competitions while working within their companies, further extending their influence. Overall, 31 (checked in April 2026) out of 387 Competition Grandmasters (approximately 8\%) are employed by these three companies, indicating that Kaggle-related experience is recognized in certain industry contexts. Tracking the professional trajectories of Kaggle Masters is more challenging, as their population is roughly six times larger than that of Grandmasters, making systematic analysis less feasible. Although titles such as Kaggle Grandmaster or Master are not a primary requirement for most professional ML roles, they can be considered important signals of expertise alongside more widely established qualifications and experience.

% \subsection{Kaggle's Role in Research}
\subsection{The impact of Kaggle on machine learning research}
\label{sec:Kaggle_research} 
MLCPs, particularly Kaggle, have frequently served as inspiration for academic research papers. This influence can be seen in various ways, whether through competition winners documenting their winning methodologies and insights, researchers utilizing competition data with final LBs serving as benchmarks for comparing newer and more advanced techniques, or simply through the use of datasets hosted on the platform that have never been part of a competition. These contributions highlight Kaggle's significant role in advancing ML research and fostering innovation in the field.

To examine the relationship between Kaggle and academic research, several major publishing platforms, along with Google Scholar, were queried using the keyword ``Kaggle'', a term with minimal ambiguity (\autoref{table:kaggle_references}). The number of yearly publications referencing Kaggle shows a consistent upward trend across all platforms. However, counts from the selected platforms are notably lower than those reported by Google Scholar, largely due to its broader coverage, including preprints and non-journal sources. Additional Google Scholar queries for ``Kaggle dataset'', ``Kaggle competition'' and ``Kaggle notebook'' returned approximately 21,900, 7,200, and 1,400 results (April 2026), respectively, underscoring the prominence of Kaggle-hosted datasets in academic research.

\begin{table*}[ht!]
\caption{Number of publications mentioning ``Kaggle''}
\label{table:kaggle_references}
\centering
\begin{threeparttable}
\begin{tabular}{l | r r r r r } 
 Year & Elsevier & S.\textsuperscript{a} Conf.\textsuperscript{b} & S.\textsuperscript{a} no Conf.\textsuperscript{b} & IEEE & GS\textsuperscript{c} \\ 
 \hline

2025 & 2.486 & 2.603 & 2.823 & 1.750 & 35.200  \\ 
2024 & 1.665 & 1.953 & 1.948 & 1.382 & 43.700 \\ 
2023 & 1.221 & 1.801 & 1.475 & 866 & 42.200    \\ 
2022 & 1.073 & 1.317 & 1.040 & 606 & 31.300    \\ 
2021 & 722 & 760 & 742 & 383 & 19.700    \\ 
2020 & 382 & 371 & 379 & 135 & 11.300    \\ 
2019 & 179 & 211 & 193 & 84 & 6.540    \\ 
2018 & 117 & 95 & 117 & 40 & 3.890   \\  
2017 & 43 & 44 & 79 & 23 & 2.080   \\ 
2016 & 28 & 17 & 34 & 10 & 1.120   \\ 
2015 & 16 & 18 & 25 & 8 & 844    \\ 
2014 & 17 & 9 & 16 & 4 & 515    \\ 
2013 & 10 & 8 & 15 & 5 & 412    \\ 
2012 & 5 & 6 & 5 & 2 & 227    \\ 
2011 & 4 & 2 & 1 & 2 & 78    \\ 
   
\hline
\end{tabular}
\begin{tablenotes}[flushleft]\footnotesize
    \item[a] S. stands for Springer
    \item[b] Conf. stands for Conference (also, about \~=95\%  of IEEE is from conferences)
    \item[c] GS stands for Google Scholar
    \end{tablenotes}
\end{threeparttable}

\end{table*}

\begin{figure}[ht!]
    \centering
    \includegraphics[width=0.75\textwidth]{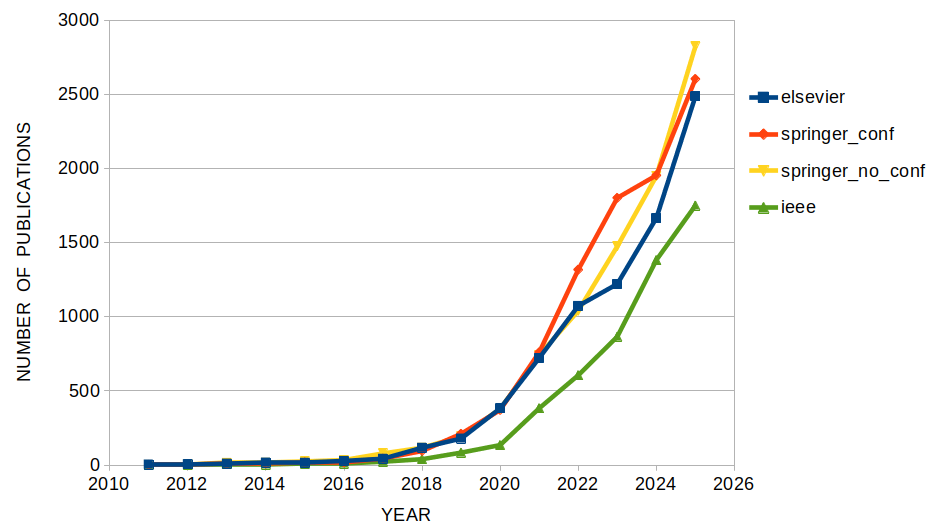}
    % {kaggle_in_journals.png}
    \caption{Annual number of Kaggle-related publications by publisher}
    \label{fig:kaggle_research_yearly}
\end{figure}

\autoref{fig:kaggle_research_yearly} illustrates the continuous and exponential rise in the use of Kaggle in academic research papers, with a particularly noticeable surge after 2020. This upward trend reflects the increasing adoption of Kaggle not only as a valuable resource for data science competitions but also as a tool for generating datasets and advancing research. %, benchmarking models, and advancing research methodologies. 
The growth is evident across various types of publications, including both peer-reviewed academic journals and conference proceedings, highlighting Kaggle's expanding influence within the research community.

\section{Impact of machine learning competitions on the AI ecosystem}
\label{sec:MLCrole}
% \section{Discussion}
% \label{sec:discussion}

The comparative analysis of MLCPs revealed their core strengths as well as the challenges they face in fostering innovation. This section examines the benefits of these competitions, highlighting their contributions relative to both academic research and industry applications.

% \subsection{Advantages of Engaging in Machine Learning Competitions}
\subsection{Advantages for participants}
\label{sec:MLCadvantages}

MLCs provide a wide range of benefits for individuals, teams, and organizations, helping to accelerate learning, innovation, and practical application in the field.

\subsubsection{Skill development}
\label{sec:skill_development}  
MLCs provide an effective environment for developing both technical and practical data science skills. Participants gain experience working with complex, incomplete, and unbalanced datasets that closely reflect real-world conditions. Through iterative experimentation, they refine their abilities in feature engineering, model tuning, and performance evaluation while learning to design solutions that generalize effectively to unseen data. Many competitions also impose computational or time constraints, encouraging efficient and scalable model development. Beyond technical expertise, team-based challenges foster collaboration, communication, and project management, allowing participants to learn from peers and practice interdisciplinary teamwork. Collectively, these experiences help bridge the gap between theoretical knowledge and applied problem-solving, preparing participants for diverse roles across academia and industry.

\subsubsection{Accelerated learning}
\label{sec:accelerated_learning}
Participating in MLCs fosters accelerated learning by providing practical experience with real-world problems that demand skills in data preprocessing, model selection, and performance evaluation. Competitors gain exposure to new algorithms, methodologies, and techniques through shared code, peer discussions, and solution reviews. These challenges also offer a fast feedback loop via live LB rankings, enabling participants to quickly iterate and refine their models based on performance metrics. This combination of hands-on practice, collaborative learning, and immediate feedback creates an intensive environment for mastering advanced data science and ML concepts.

\subsubsection{Exposure to real-world problems}
\label{sec:real-world_problems}
MLCs offer participants exposure to diverse applications across domains like healthcare, finance, image recognition, NLP, and recommendation systems, broadening their understanding of various industries. They also tackle challenging, cutting-edge, or unsolved problems, encouraging creative thinking and innovative approaches beyond conventional textbook scenarios. This exposure equips competitors with a practical perspective on how ML is applied to solve complex, impactful issues in real-world contexts.    

\subsubsection{Networking and community building}
\label{sec:networking}
MLCs foster networking and community building by connecting participants with top-tier experts and practitioners worldwide, creating opportunities for collaboration, mentorship, and learning. The culture of open knowledge sharing allows competitors to exchange ideas and contribute to open-source knowledge by sharing solutions and methodologies, advancing the field collectively.

\subsubsection{Career advancement}
\label{sec:career}  
Participating in MLCs can boost career advancement by helping individuals build a strong portfolio showcasing their practical skills and accomplishments, which appeals to prospective employers and collaborators. In some cases, platforms hosting these competitions are used by companies and organizations to identify and engage with talented participants, although systematic evidence on the extent of this practice remains limited. High-performing participants can also gain visibility through rankings, awards, and public recognition, which may support professional opportunities.

\subsubsection{Access to high-quality datasets}
\label{sec:data_access}
Competitions offer participants access to high-quality, often proprietary datasets that would otherwise be unavailable, supporting the development and evaluation of advanced models. These datasets frequently contain domain-specific information, such as in medical imaging or financial transactions, allowing participants to acquire specialized expertise and apply their skills to targeted fields.

\subsection{Benefits for industry and academia}
\label{sec:benefits_Ind_Ac}
The advantages of MLCs extend beyond participant advantages, generating tangible value for both industry and academia.

\subsubsection{Leveraging participant expertise in industry and academia}
\label{sec:participant2AcInd}
Organizations and research institutions directly benefit from this continuous upskilling. Companies gain employees and collaborators capable of designing, optimizing, and deploying advanced models efficiently, while academic environments profit from researchers who incorporate competition-driven methodologies into scientific inquiry and pedagogy. 

\subsubsection{Innovation and research advancement}
\label{sec:innovation}
Competitions foster creative solutions by encouraging participants to develop novel approaches to challenging problems, often leading to significant research insights or breakthroughs. They serve as platforms for benchmarking new techniques against existing methods, driving the improvement of models and algorithms. Furthermore, the exposure to diverse domains and methods promotes the cross-pollination of ideas, where innovations in one field inspire advancements in another.

\subsubsection{Crowdsourcing solutions for companies}
\label{sec:crowd_solutions}
MLCs provide companies with access to a diverse global talent pool, enabling them to crowdsource innovative solutions that may not emerge from internal teams alone. This approach serves as a cost-effective alternative to traditional research and development, allowing organizations to tackle complex problems without significant upfront investment. Additionally, hosting competitions fosters community engagement, helping companies connect with the ML community, attract top-tier talent, and spotlight their technology or challenges, ultimately strengthening their industry presence.

\subsection{Positioning MLCs within the broader AI ecosystem}
\label{sec:MLC_thBpr}
% From Research to Real-World Impact: The Place of ML Competitions

ML advances through multiple pathways as is the traditional academic research, industrial applications, MLCs and other contributors. Each channel plays a unique role in driving innovation, shaping technology, and solving real-world problems. A structured analysis is presented, examining how ML progresses through these avenues, highlighting their differences and complementary roles.

Academia has long been the principal driver of theoretical and foundational progress in ML. Academic research has produced many of the mathematical and statistical underpinnings of the field, with landmark innovations such as backpropagation, support vector machines, gradient boosting, and transformer architectures originating in academic settings. Universities and research institutions provide the intellectual freedom to explore long-term, unsolved challenges without the constraints of commercial imperatives. Moreover, the peer-review process in leading venues such as NeurIPS, ICML, and CVPR ensures scientific rigor, reproducibility, and the continuous refinement of ideas that collectively advance the discipline.

In contrast, industry-led ML research is driven primarily by application and scale. Companies leverage ML to address real-world problems, from recommendation systems and fraud detection to speech recognition and autonomous systems, while simultaneously advancing the state of applied AI. Corporations benefit from access to massive datasets and extensive computational infrastructure, enabling the training of models at unprecedented scales. Industrial research focuses on operationalizing ML systems, emphasizing scalability, robustness, and real-time performance. Furthermore, interdisciplinary collaboration among engineers, researchers, and domain specialists fosters the creation of practical tools and frameworks such as TensorFlow and PyTorch, which, in turn, feed back into academic and open-source research. Together, academic inquiry and industrial innovation form a synergistic cycle that continually propels the development and adoption of ML technologies.

\subsubsection{Contrasting ML practices in competitions, academia, and industry}
\label{ComparingApproaches}   

A comparative overview of their roles, strengths, and differences is provided in \autoref{table:AiAcInd}. MLCs, academia, and industry each contribute to the advancement of ML, but they do so with distinct motivations and often drive progress in different directions. MLCs focus on practical problem-solving, fostering innovation through competition-driven challenges. Academia, on the other hand, emphasizes theoretical advancements, developing new methodologies and pushing the boundaries of foundational research. Meanwhile, industry applies ML at scale, prioritizing efficiency, commercialization, and real-world impact. MLCs often operate at the intersection of academia and industry, acting both as a bridge between the two and as an independent force driving innovation. These three domains interact, influencing one another in various ways, yet their core objectives remain unique.

\begin{center}
\captionof{table}{Comparison: Academia vs. MLCs vs Industry}
\label{table:AiAcInd}
\begin{tabular}{|p{6em}|p{8.5em}|p{8.5em}|p{8.5em}|}

  \hline
\textbf{Factor} & \textbf{Academia} & \textbf{MLCs} & \textbf{Industry} \\
 \hline

Focus & 
Theoretical advancements, fundamental research & 
Practical problem-solving and benchmarking & 
Solving real-world business problems, productizing ML \\

\hline

Innovation Drivers  & 
Long-term, unsolved challenges & 
Crowdsourced, rapid experimentation & 
Application-driven, scale and efficiency \\

\hline

Data Access & 
% Usually smaller, specific to research area 
Smaller, domain-specific  or larger, aggregated from multiple sources.& 
Often public, curated datasets & 
Access to massive, proprietary datasets \\

\hline

Scalability & 
Typically less emphasis on scalability & 
Maximizing performance on a fixed dataset & 
Emphasis on deploying models at scale \\

\hline

Learning and Collaboration & 
Collaboration through conferences and journals & 
Open collaboration, community-driven & 
Interdisciplinary teams, proprietary innovations \\

\hline

Time Horizon & 
Long-term, theoretical work & 
Short-term, rapid iteration & 
Immediate to mid-term product development \\

\hline

Recognition & 
Peer-reviewed publications & 
Leaderboards, prizes & 
Patents, product improvements, business outcomes \\

  \hline
\end{tabular}

\end{center}

\subsubsection{Interactions between MLCs and external ML contributors}
\label{sec:other_contr}

Beyond academia, industry, and dedicated competition platforms, a variety of actors play crucial roles in shaping the MLC ecosystem. Open-source communities such as TensorFlow, PyTorch, Scikit-learn \citep{pedregosa2011scikit}, and Hugging Face \citep{huggingface} provide the tools, frameworks, and pre-trained models that participants frequently leverage to develop competitive solutions, enabling rapid experimentation and reproducibility while, under the influence of MLCs, gaining advantages by refining or extending their codes. Independent researchers and hobbyists further enrich competitions by contributing innovative approaches, sharing code repositories, and disseminating novel strategies that often influence LB outcomes and post-competition publications while being acknowledged for their contributions. Governments and public institutions support these activities indirectly by funding AI research, establishing strategic priorities, and promoting data sharing initiatives that supply high-quality datasets for competitions in crucial domains such as healthcare, environmental monitoring, and public safety.

Non-profit organizations, including the Allen Institute for AI, OpenAI in its early open-research phase, and the Mozilla Foundation, contribute by advocating for ethical and transparent AI practices, which guide competition design and responsible dataset usage. Science communicators and media platforms broaden awareness of competitions and their societal relevance, attracting diverse participants and fostering community engagement. Finally, policymakers and ethics boards shape the regulatory frameworks that govern data privacy, fairness, and accountability within competitions.

Together, these contributors augment the efforts of academia, industry, and competition hosts, creating a dynamic ecosystem that fosters innovation, inclusivity, and ethical standards in MLCs. In turn, they benefit from these competitions by accessing cutting-edge research, engaging with a skilled and diverse community, and using competition outcomes to guide their initiatives, advocacy, and policy development.

\subsection{Evolving trends in machine learning competitions}
\label{sec:MLC_future}
MLCs are expected to play an even greater role in the future of AI research and development as the number of participants continues to grow. This trend is further reinforced by the growing incorporation of MLC outcomes into academic research. As demonstrated by the growing presence of Kaggle-related studies in academic journals and conferences in \autoref{sec:Kaggle_research}, MLCPs are becoming valuable tools for benchmarking algorithms and evaluating new methodologies. Universities may incorporate MLCs more extensively into their curricula, using them not only as learning exercises but also as controlled settings for experimenting with and optimizing research methodologies. 

Industry collaboration is also expected to expand, with more organizations leveraging competitions to solve real-world problems across various domains, such as healthcare, finance, and defense. Instead of purely theoretical challenges, future competitions may focus on applied problems, requiring interdisciplinary knowledge and domain expertise. This shift could create stronger connections between competition winners and industry stakeholders, offering more opportunities for direct employment, startup funding, or long-term research collaborations.

As ML advances, competitions could extend traditional LB-driven approaches by incorporating evaluation metrics that capture model robustness, interpretability, and fairness. Rather than replacing LBs, these aspects could be reflected through composite or multi-objective scoring schemes that better align performance with broader real-world requirements. The increasing popularity of AutoML and low-code platforms may reduce the emphasis on hyperparameter tuning and model optimization, shifting focus toward problem formulation, data quality, and ethical considerations. Instead of simply rewarding the most accurate predictions, future competitions might emphasize long-term performance, adaptability to unseen conditions, and responsible AI principles. Similarly, continuous benchmarking initiatives could replace time-limited contests, encouraging researchers to iteratively improve solutions over time.

\section{Limitations and future research}
\label{sec:limitation_futurework}
Despite providing a broad overview of MLCs and their role in advancing the field, this study has several limitations. Parts of the analysis rely on publicly available data from competition platforms and third-party analytics, which may be incomplete, subject to updates, or influenced by short-term fluctuations. Additionally, the diversity and rapid evolution of MLCPs make it difficult to capture all competition formats and emerging trends within a single study.

A key limitation of MLCs concerns the extent to which competition outcomes reflect generalizable progress. Competitive settings often incentivize optimization for LB performance, which can encourage specialization to specific datasets or sub-populations at the expense of robustness. Moreover, many competitions report single-point performance metrics without statistical significance testing or uncertainty estimates, meaning that small, potentially insignificant differences can determine final rankings. Evaluation criteria are also typically narrow, focusing on predictive accuracy while overlooking aspects such as fairness, interpretability, energy consumption, and broader societal impact. The substantial computational resources required for top performance may further introduce barriers to participation and environmental costs, while potentially shifting focus toward incremental metric gains.

Future research can address these limitations in several ways. Longitudinal analyses could better capture the evolution of competition topics, participation, and methods over time, while comparisons between competition results and academic research may clarify knowledge transfer and generalization. Further work is needed to assess the trade-offs between LB optimization and robustness, including the use of statistical testing and multi-dimensional evaluation metrics. Expanding competition frameworks to incorporate fairness, efficiency, and societal impact would support more holistic benchmarking. Finally, investigating the educational, ethical, and economic implications of MLCs would contribute to a more comprehensive understanding of their role in the broader AI ecosystem.

\section{Conclusions}
\label{sec:Conclusions}
MLCs have emerged as powerful catalysts for AI advancement, effectively connecting academic research, industrial applications, and real-world problem-solving. This study examined major MLCPs, analyzing their structures, participation patterns, and overall contributions to the AI ecosystem. MLCs continue to drive innovation across academia and industry, with top-performing solutions frequently inspiring subsequent research or leading to deployable, production-ready technologies. The concurrent evolution of open-source frameworks, governmental initiatives, and ethical AI discourse is further influencing competition design, evaluation standards, and participation models.

Kaggle remains the dominant MLCP, although several newer platforms have gained significant traction in recent years. Their collective strength lies in the vitality of their data science communities, which share common goals yet differ in workflow design, evaluation methodologies, and engagement strategies. Competition quality can be evaluated using several indicators, including the number of participants, their experience level (as reflected in Kaggle's tier system or prior winning placements), the credibility of the host, and the platform's audience reach. Analysis of Kaggle's metadata revealed that leading participants predominantly originate from Japan, United States, European Union, and China, while India's rapidly expanding AI community is becoming increasingly prominent. Some high-performing competitors transition into roles at leading AI firms, suggesting that MLC participation can contribute to career development in the ML field. %Many high-performing competitors transition into positions at leading AI firms, emphasizing the pivotal role of MLCs in career development within the ML field. 
Furthermore, the growing number of research publications citing Kaggle, underscores its expanding academic influence. Results derived from competitions, strengthened through repeated validation and collective peer scrutiny, provide a more reliable basis for future research. The diversity of competition hosts, including corporations, universities, non-profit organizations, and government agencies, illustrates the broad and inclusive impact of these platforms.

MLCs offer valuable opportunities to participants to refine technical expertise, apply skills to real-world problems, and advance professional trajectories. Industry benefits from these ecosystems through access to innovative solutions, efficient talent identification, and enhanced brand visibility. Academic institutions, in turn, gain from reproducible benchmarks, collaborative networks, and new research directions. Collectively, competitions foster rapid experimentation, knowledge sharing, and interdisciplinary collaboration, bringing together participants, independent researchers, and professionals from across sectors.

As global hubs for AI innovation, MLCs leverage open-source contributions, community engagement, and cross-sector partnerships to produce practical, high-impact solutions. Their roles in talent discovery, research advancement, and industrial adoption are poised to expand further. However, ensuring fairness, inclusivity, and ethical alignment remains critical for sustaining their long-term positive impact. By fostering open collaboration, reproducible research, and merit-based recognition, MLCs will continue to shape the next generation of AI practitioners and accelerate technological breakthroughs across multiple domains.

\section*{Acknowledgements}
% \bmhead{Acknowledgements}
I would like to thank the reviewers for the insightful commentary and suggestions, which improved the quality of this manuscript. 

\section*{Declarations}
\begin{itemize}
\item \textbf{Funding}
No funding was received to assist with the preparation of this manuscript.
\item \textbf{Competing interests} 
Author has no relevant financial or non-financial interests to disclose.
% The author declares that has no known competing financial interests or personal relationships that could have appeared to influence the work reported in this paper.
% \item Ethics approval and consent to participate
% Not applicable
% \item Consent for publication
% Not applicable
\item \textbf{Data availability} 
The dataset that was used for kaggle users statistics is available at \href{https://www.kaggle.com/datasets/kaggle/meta-kaggle}{Meta Kaggle}.
% \item Materials availability
% Not applicable
\item \textbf{Code availability} 
Author's public notebook \url{https://github.com/IoannisNasios/various/blob/master/c_m_gm_g_h_.ipynb}

% \href{https://www.kaggle.com/code/ouranos/c-m-gm-g-h}{https://www.kaggle.com/code/ouranos/c-m-gm-g-h}
% % \item Author contribution
% % Not applicable
\end{itemize}

%\bibliographystyle{plain} % We choose the "plain" reference style
% \bibliographystyle{cas-model2-names}
%\bibliographystyle{abbrvnat}


\bibliography{refs}

\end{document}